\documentclass[twocolumn]{article}

\usepackage[utf8]{inputenc}
\usepackage[margin=1.25in]{geometry}
\usepackage{authblk}
\usepackage{setspace}
\usepackage{graphicx}
\graphicspath{{./figures/}}
\usepackage{subcaption}
\usepackage{multirow}
\usepackage{amsmath,amssymb,amsfonts}
\usepackage{amsthm}
\usepackage{mathrsfs}
\usepackage{xcolor}
\usepackage{textcomp}
\usepackage{booktabs}
\usepackage{array}
\usepackage{tabularx}
\usepackage{hyperref}

\usepackage[switch]{lineno}
\newcommand{\standardtablesetup}{%
  \footnotesize
  \setlength{\tabcolsep}{2.6pt}%
  \renewcommand{\arraystretch}{1.18}%
}
\usepackage[
  style=nejm,
  citestyle=numeric-comp,
  sorting=none
]{biblatex}
\title{Edge-Constrained UAV Small-Object Detection with P2 Enhancement and Quantum-Inspired Lightweight Structure Search}

\author[1,*]{Wuming Lei}
\author[1]{Yanbin Gao}
\author[1]{Mingyan Sun}
\author[1]{Xiaobin Li}
\author[1]{Xuechen Liang}
\affil[1]{East China Jiaotong University, Nanchang, China.}
\affil[*]{Corresponding author: Wuming Lei; email: \href{mailto:WMFCDS@outlook.com}{WMFCDS@outlook.com}}

\date{}

\begin{document}
\twocolumn[
\begin{@twocolumnfalse}

\maketitle

\begin{abstract}
Unmanned aerial vehicle (UAV) object detection requires compact detectors that retain small-object details under onboard computation and memory constraints. Repeated downsampling in lightweight networks weakens shallow spatial information, while manually adding attention or fusion modules may increase cost without stable gains. This study analyzes YOLOX-Nano under edge-deployment constraints by combining a P2 high-resolution detection branch with a quantum-inspired evolutionary algorithm (QIEA) for lightweight structure screening. The search space is defined by lightweight priority and task specificity, and the evaluation jointly considers accuracy, floating-point operations (FLOPs), latency, memory consumption, and recall. On VisDrone, the P2 branch increases AP$_{\mathrm{small}}$ by 31.10\% over the YOLOX-Nano baseline. Compared with NanoDet-Plus with similar model size, YOLOX-Nano+P2 improves AP$_{50:95}$ by 17.5\% and AP$_{\mathrm{small}}$ by 44.9\%. The QIEA-selected candidate obtains the highest Recall$_{50}$, but +P2 remains the strongest AP-oriented variant after full training. Full 100-epoch verification of Random-best, GA-best, and SA/QUBO-best candidates further shows that proxy rankings do not necessarily transfer to final AP$_{50:95}$. These results support using P2 as the main small-object enhancement path and QIEA as a lightweight tool for candidate screening and accuracy-cost analysis. The source code, configuration files, diagnostic scripts, and summarized results are available at \url{https://github.com/Ming23233/UAV-QIEA-Edge-Detection}.
\end{abstract}

\vspace{0.75em}
\noindent\textbf{Keywords:} UAV object detection; small-object detection; edge computing; YOLOX-Nano; P2 high-resolution branch; quantum-inspired evolutionary algorithm; lightweight structure search
\vspace{1em}

\end{@twocolumnfalse}
]

\section{Introduction}\label{sec:introduction}

Object detection in UAV aerial imagery has become a core component of low-altitude intelligent perception. Public benchmarks such as VisDrone, UAVDT, and DOTA have provided standardized evaluation protocols for aerial, traffic, and remote-sensing detection tasks \cite{zhu2018visdronechallenge,du2018uavdt,xia2018dota}. The VisDrone challenge further reveals that dense small objects, large scale variation, severe occlusion, and class imbalance are persistent difficulties in UAV imagery \cite{du2021visdrone}. Recent surveys on UAV and remote-sensing object detection also indicate that scale variation, dense object distribution, and complex backgrounds remain major causes of missed detections, false positives, and localization errors \cite{hu2024uavsurvey,cheng2016remote,zhu2017deepremote}.

From the perspective of detection frameworks, two-stage detectors, one-stage detectors, feature pyramids, and dense detection losses have provided the technical basis for subsequent UAV detection systems \cite{lin2017fpn,lin2017focal,tian2019fcos,liu2020generic}. Faster R-CNN, SSD, Libra R-CNN, and ATSS represent region-proposal detection, one-stage dense detection, balanced detection learning, and adaptive sample assignment, respectively \cite{ren2017faster,liu2016ssd,pang2019libra,zhang2020atss}. YOLOX improves real-time detection through a decoupled head, an anchor-free design, and an optimized training strategy, which makes YOLOX-Nano a suitable lightweight baseline for controlled analysis \cite{ge2021yolox}. In addition, tracking-by-detection systems such as ByteTrack show that reliable detection boxes are essential for downstream UAV perception pipelines \cite{zhang2022bytetrack}.

UAV platforms and edge devices are usually constrained by computation, memory, power consumption, and response time. Edge intelligence studies emphasize that moving artificial intelligence tasks from cloud servers to edge platforms requires joint consideration of accuracy, latency, bandwidth, and energy \cite{zhou2019edge}. Efficient deep learning and model-compression studies also suggest that deployable detectors should be evaluated not only by accuracy but also by parameters, FLOPs, inference latency, and memory footprint \cite{chen2019edgeDL,cheng2018compression,menghani2023efficient}. Therefore, UAV edge detection requires a deployment-oriented analysis that links recognition performance with hardware cost.

Existing UAV small-object detection methods have investigated clustered regions, feature enhancement, and multi-scale fusion \cite{yang2019clustered,wan2021vistrongerdet}. Recent YOLO-based studies further improve UAV small-object detection by introducing lightweight modules, attention mechanisms, dynamic heads, and multi-scale fusion \cite{xu2024yolov8n,zhao2024tayolo,liu2020uavyolo,wang2023improvedyolox,wang2024smalltargetyolov5,liu2024sodyolo,chen2024hspyolov8,luo2025enhanced}. Nevertheless, three limitations remain. First, computation reduction in lightweight models may weaken shallow spatial details, thereby affecting small-object localization and recall. Second, manually stacking attention, context, or fusion modules does not always bring stable gains because their effects depend on object-scale distribution and model capacity. Third, many studies focus mainly on accuracy and give limited attention to parameters, FLOPs, latency, and memory consumption.

This study analyzes UAV small-object detection under edge-computing constraints using YOLOX-Nano as the controlled base detector. The analysis connects three components: P2 high-resolution feature enhancement, QIEA-inspired candidate screening, and marginal accuracy-cost evaluation. It focuses on structure-level evidence within YOLOX-Nano rather than on cross-framework ranking, and places high-resolution enhancement, low-budget probabilistic screening, and deployment cost in the same decision space.

The main contributions are summarized as follows:
\begin{itemize}
    \item First, we build an edge-constrained UAV small-object detection analysis framework around YOLOX-Nano and evaluate P2 high-resolution enhancement together with parameters, FLOPs, latency, and memory consumption.
    \item Second, we formulate a QIEA-inspired lightweight candidate-screening process with Q-bit probability encoding, categorical probability vectors, explicit update equations, and a complexity-penalized proxy fitness function.
    \item Third, we provide an evidence chain covering multi-seed experiments, lightweight method comparison, ablation analysis, heuristic search comparison, full-training verification, efficiency evaluation, AU-AIR external testing, and small-object diagnostic analysis.
\end{itemize}

\begin{figure*}[t]
    \centering
    \includegraphics[width=\textwidth]{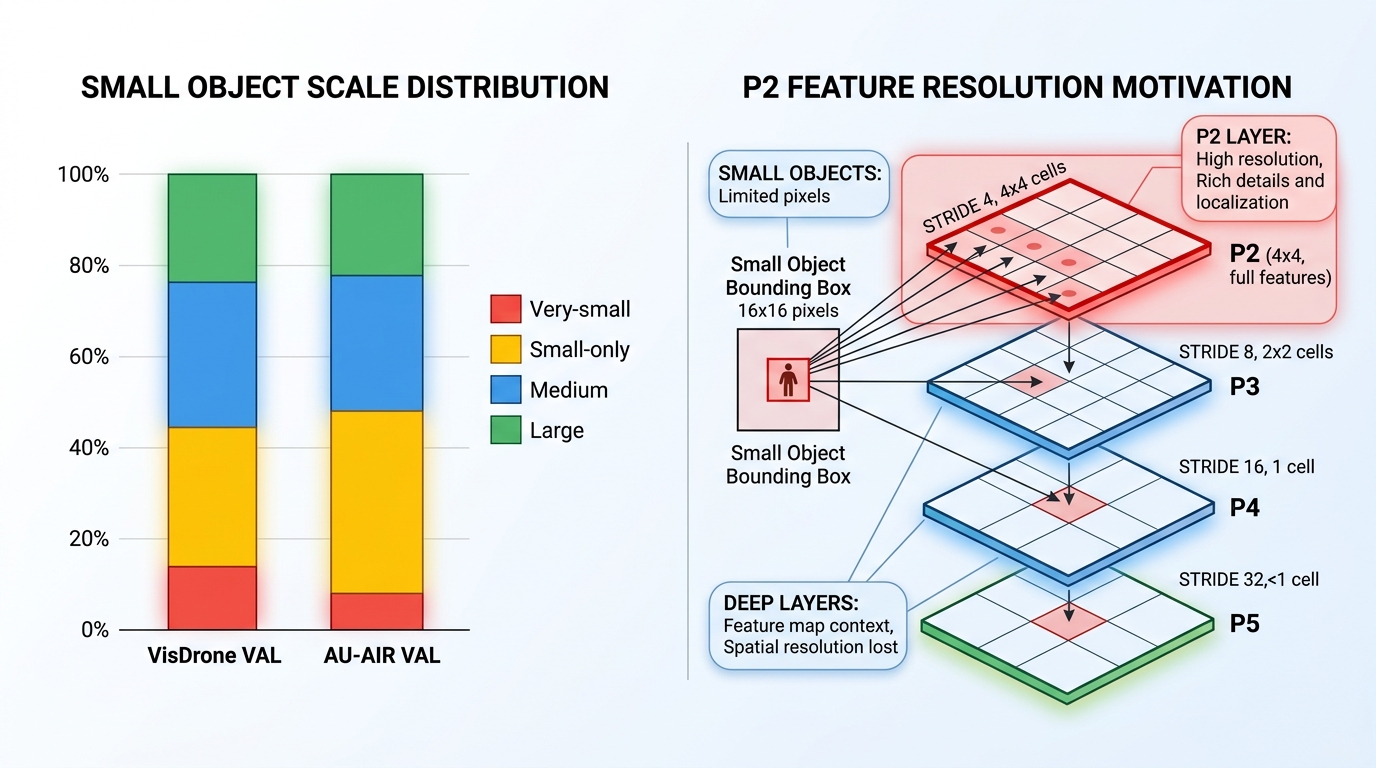}
    \caption{Small-object scale distribution and feature-resolution motivation. The left panel compares object-scale distributions on VisDrone and AU-AIR validation splits, while the right panel illustrates how the P2 feature level preserves higher spatial resolution for a $16\times16$ UAV target than deeper feature levels.}
    \label{fig:motivation-scale-p2}
\end{figure*}

To make the small-object motivation explicit, Fig.~\ref{fig:motivation-scale-p2} summarizes the object-scale distribution and the high-resolution feature requirement. VisDrone contains a much larger proportion of very-small and COCO-small objects than AU-AIR, indicating that UAV detection results are strongly affected by object-scale distribution. For a representative $16\times16$ target, the stride-4 P2 feature level preserves about $4\times4$ spatial samples, whereas deeper feature levels rapidly reduce the target to only a few samples or less than one effective sample. This observation supports evaluating the P2 branch before screening additional lightweight structures. The related work below reviews the three design factors involved in this choice: high-resolution features, lightweight deployment, and search-based structure selection.

\section{Related work}\label{sec:related}

\subsection{UAV small-object detection}\label{subsec:uav-small}

UAV object detection research usually focuses on scale variation, dense objects, occlusion, and complex backgrounds. In addition to the benchmarks introduced above, DroneVehicle provides another UAV-based vehicle detection and counting dataset \cite{sun2022dronevehicle}. Feature Pyramid Networks and Focal Loss, introduced in the general detection literature, motivate multi-scale representation and foreground-background balancing for small targets. Studies on UAV detection, remote-sensing detection, and small-object detection show that small targets rely heavily on high-resolution spatial information, contextual cues, and cross-scale feature fusion \cite{tong2020small}. High-resolution representation studies such as HRNet also show that preserving fine spatial information is important for visual recognition tasks \cite{sun2019hrnetpose,wang2021hrnet}. Therefore, adding a high-resolution branch to a lightweight detector has a clear task motivation.

In aerial scenes, the previously cited clustered-region and VisDrone-enhancement studies show the importance of dense-region handling and visual information enhancement, and multi-scale feature fusion remains a common way to recover small-target information \cite{li2020multiscale}. Recent YOLO-based UAV methods further confirm that single-scale or low-resolution outputs are insufficient for UAV small-object detection. Unlike these detector-improvement studies, this work uses YOLOX-Nano as a controlled base model and focuses on the relationship among high-resolution enhancement, lightweight candidate screening, and deployment cost.

\subsection{Lightweight detection and edge deployment}\label{subsec:edge}

Lightweight model design commonly reduces computation with depthwise separable convolution, inverted residuals, compound scaling, and efficient detection heads. MobileNet and MobileNetV2 establish important mobile convolutional design principles, while EfficientNet and EfficientDet discuss scaling and detection efficiency \cite{howard2017mobilenets,sandler2018mobilenetv2,tan2019efficientnet,tan2020efficientdet}. Model compression, pruning, quantization, automatic compression, and efficient deep learning also provide important tools for edge deployment \cite{han2016deepcompression,he2018amc}.

Based on the edge-intelligence motivation introduced in the Introduction, this work reports parameters, FLOPs, GPU batch-1 latency, and peak memory consumption to analyze the deployment cost corresponding to accuracy improvement. This places the paper closer to edge-oriented performance evaluation than to a pure detection-accuracy comparison.

\subsection{Structure search and quantum-inspired optimization}\label{subsec:search}

Neural architecture search reduces manual design effort by automatically searching for network structures. NAS surveys summarize search space, search strategy, and performance estimation \cite{elsken2019nas}, while neuroevolution provides another line of population-based neural structure optimization \cite{stanley2019neuroevolution}. ProxylessNAS, Once-for-All, and MnasNet further connect search with target tasks or hardware constraints \cite{cai2019proxylessnas,cai2020ofa,tan2019mnasnet}. DARTS and FBNet show that architecture search can support efficient network design and mobile deployment \cite{liu2019darts,wu2019fbnet}. NAS-FPN directly links feature pyramid design with architecture search for detection \cite{ghiasi2019nasfpn}.

Quantum-inspired evolutionary algorithms represent combinatorial variables by probability amplitudes and update the search distribution according to fitness feedback. The QEA of Han and Kim uses Q-bit individuals and quantum gate updating for combinatorial optimization \cite{han2002qea}. Related studies in feature selection, quantum-inspired metaheuristics, electromagnetic device design, QIEA-NAS, and QUBO modeling demonstrate the feasibility of quantum-inspired or quantum-related optimization for discrete variable search \cite{vivek2024qieasurvey,muecke2023feature,zhang2012qiea,pooja2024scientometric,szwarcman2022qiea,glover2019qubo}. This study adopts the Q-bit probability encoding and elite-guided distribution update ideas for detector structure screening, while the actual implementation remains a classical proxy-search procedure.

\subsection{Attention and context modules}\label{subsec:attention}

Attention and context modeling are important for improving feature representation in detection. SE recalibrates channels, Coordinate Attention introduces positional information into lightweight attention, and CBAM combines channel and spatial attention \cite{hu2018senet,hou2021coordinate,woo2018cbam}. Deformable Convolution and DetectoRS improve detection representation through spatial adaptation and recursive feature pyramids \cite{dai2017deformable,qiao2021detectors}. PANet, DETR, and Swin Transformer further show that path aggregation, end-to-end detection, and hierarchical vision backbones affect detection features \cite{liu2018panet,carion2020detr,liu2021swin}. These modules provide useful design directions, but adding all of them to a nano-scale detector may increase optimization difficulty and deployment cost. The search space is therefore kept compact and guided by lightweight priority, small-object specificity, and controllable comparison.

\section{Materials and Methods}\label{sec:method}

\subsection{Overall framework}\label{subsec:framework}

The proposed workflow is built on YOLOX-Nano and follows a high-resolution enhancement, lightweight structure search, and deployment evaluation pipeline. The main dataset is first converted to COCO format, and its object-scale distribution is analyzed to confirm the dominance of small objects. A YOLOX-Nano baseline is then trained, and a P2 high-resolution detection branch is introduced into the feature pyramid to enhance shallow spatial details. On this basis, attention, context, scale fusion, small-object loss weight, and center sampling radius are included in a discrete search space. A QIEA-inspired proxy search evaluates candidate structures under a limited budget. The selected models are verified through multi-seed training, ablation analysis, search comparison, efficiency evaluation, and an AU-AIR external engineering case.

\begin{figure*}[t]
    \centering
    \includegraphics[width=\textwidth]{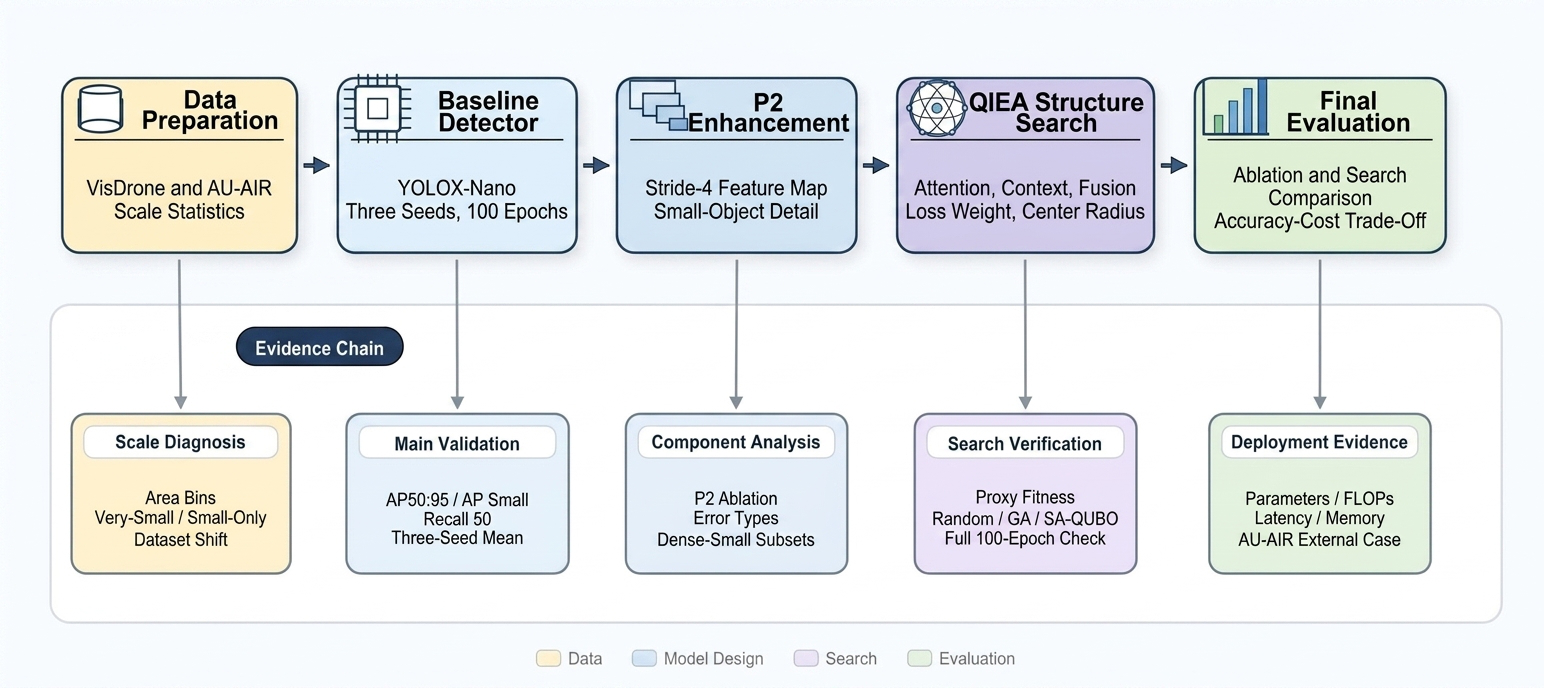}
    \caption{Experimental workflow and evidence chain. The main path proceeds from data preparation and YOLOX-Nano baseline training to P2 high-resolution enhancement, QIEA structure search, and final evaluation. The evidence chain aligns each stage with the corresponding diagnostics, including scale analysis, three-seed validation, component ablation, search verification, deployment-cost measurement, and AU-AIR external evaluation.}
    \label{fig:workflow}
\end{figure*}

Figure~\ref{fig:workflow} summarizes the workflow as a main experimental path and an aligned evidence chain. Scale diagnosis supports the dataset motivation, three-seed validation and component analysis verify model behavior, and search verification and deployment evidence quantify proxy stability and edge-oriented feasibility. This separation is important because an architecture that improves proxy fitness may not necessarily improve final detection accuracy after full training. The following sections define the detection objective, the P2 branch, and the QIEA-inspired search procedure before reporting the experimental results.

\subsection{Detection and search objective}\label{subsec:objective}

Given a training set $D=\{(I_j,Y_j)\}$, the detection model $f_\theta$ is trained by minimizing the classification, localization, and objectness losses:
\begin{equation}
    \mathcal{L} = \mathcal{L}_{\mathrm{cls}} + \mathcal{L}_{\mathrm{box}} + \mathcal{L}_{\mathrm{obj}} .
    \label{eq:det-loss}
\end{equation}
For each input image $I$, the detector predicts bounding boxes $B=\{b_i\}$, class labels $c_i$, and confidence scores $s_i$, where a bounding box is denoted as
\begin{equation}
    b_i = (x_i,y_i,w_i,h_i).
    \label{eq:bbox}
\end{equation}
For UAV scenes, small objects occupy only a limited number of pixels. The P2 branch introduces a higher-resolution feature map $F_{P2}$ so that small-scale targets retain more effective spatial responses before excessive downsampling.

The evaluation metrics include AP$_{50:95}$, AP$_{50}$, AP$_{\mathrm{small}}$, and Recall$_{50}$. Following the COCO-style area ranges, AP$_{\mathrm{small}}$ is computed for objects whose bounding-box area is smaller than $32^2$ pixels, AP$_{\mathrm{medium}}$ for objects in $[32^2,96^2)$ pixels, and AP$_{\mathrm{large}}$ for objects no smaller than $96^2$ pixels. For the additional diagnostic analysis, very-small objects are defined by bounding-box area smaller than $16^2$ pixels. Average precision can be expressed as
\begin{equation}
    \mathrm{AP} = \int_0^1 p(r)\,dr,
    \label{eq:ap}
\end{equation}
where $p(r)$ is the precision--recall curve. The inference speed can be approximated by
\begin{equation}
    \mathrm{FPS} \approx \frac{1000}{\mathrm{Latency(ms)}} .
    \label{eq:fps}
\end{equation}

In the structure search stage, a candidate architecture $a$ is sampled from a discrete search space $A$. Each candidate is evaluated by 10-epoch proxy training. The proxy fitness is defined as
\begin{equation}
    \begin{aligned}
    \mathrm{Fitness}(a) ={}&
    \mathrm{AP}_{\mathrm{small}}
    + 0.30\,\mathrm{AP}_{50} \\
    &+ 0.20\,\mathrm{Recall}_{50}
    - C(a),
    \end{aligned}
    \label{eq:fitness}
\end{equation}
where AP$_{50:95}$ is used as a fallback when AP$_{\mathrm{small}}$ is unavailable in the proxy summary. The complexity penalty is
\begin{equation}
    \begin{aligned}
    C(a) ={}& 0.0015m
    + 0.002\max(w_s-1,0) \\
    &+ 0.001|r_c-2.5|,
    \end{aligned}
    \label{eq:complexity}
\end{equation}
where $m$ is the number of enabled binary modules among CA, CSA, Context, and Fusion; $w_s$ is the small-object loss weight; and $r_c$ is the center sampling radius. The proxy objective uses AP$_{\mathrm{small}}$ as the dominant term because dense small targets are the main application focus. AP$_{50}$ and Recall$_{50}$ are included as auxiliary terms to retain localization tolerance and recall behavior during short-cycle proxy training. The complexity penalty discourages unconstrained module stacking and overly aggressive training hyperparameters. All coefficients are fixed before search and are used consistently for QIEA, random search, GA, and SA/QUBO heuristic comparison. They are not tuned on the test set and are used as a unified empirical proxy-screening criterion rather than as a learned optimal objective.

\subsection{P2 high-resolution small-object enhancement}\label{subsec:p2}

The P2 branch is designed to preserve high-resolution shallow features so that small targets retain more edge, texture, and localization information. The original YOLOX-Nano uses \texttt{dark3}, \texttt{dark4}, and \texttt{dark5} to form stride-8, stride-16, and stride-32 feature outputs. When \texttt{use\_p2=True}, this work additionally introduces the \texttt{dark2} feature from CSPDarknet and extends the neck input channels from $[256,512,1024]$ to $[128,256,512,1024]$. In YOLOPAFPN, the P2 path first reduces the higher-level feature by \texttt{reduce\_conv2}, upsamples it to the spatial resolution of \texttt{dark2}, concatenates it with the shallow feature, and then applies \texttt{C3\_p2} to produce a stride-4 feature map. This stride-4 output is appended to the neck outputs before the detection head. The detection head is correspondingly extended from three scales to four scales, and the stride list is changed from $[8,16,32]$ to $[4,8,16,32]$.

Each scale has its own stem, classification branch, regression branch, and objectness branch. Therefore, the P2 head adds a complete prediction branch with its own convolutional parameters alongside P3--P5. Except for the searched center radius, the original YOLOX dynamic label-assignment procedure is kept unchanged. This design directly targets the loss of spatial samples caused by downsampling, but it also increases FLOPs and memory consumption. Therefore, additional structures must be screened under lightweight and deployment-aware principles rather than added indiscriminately.

\subsection{QIEA-inspired lightweight structure search}\label{subsec:qiea}

The QIEA-inspired search space fixes the P2 branch as enabled and searches over CA, CSA, lightweight context modeling, scale-aware fusion, small-object loss weight, and center sampling radius. The search space is designed according to three principles. First, lightweight priority restricts the variables to modules or hyperparameters that add limited parameters and can be inserted into a nano-scale detector. Second, small-object specificity selects variables that directly affect shallow feature recalibration, local context, cross-scale information, foreground weighting, or sample assignment for small targets. Third, controllable comparability keeps the number of variables small enough for a 16-candidate proxy budget, so that QIEA, random search, GA, and SA/QUBO can be compared under the same effective search cost.

These principles explain why the search does not include backbone replacement, large transformer blocks, heavy detection heads, or full-resolution global fusion. Such options may improve accuracy in larger detectors, but they would violate the edge-constrained setting and make the proxy search budget insufficient. The selected variables cover three compact design categories: representation recalibration, local semantic enhancement, and training-signal adjustment. Thus, the search problem is not whether P2 should be used, but how to screen lightweight structures after P2 has been selected as the high-resolution enhancement path. Table~\ref{tab:qiea-space} summarizes the search variables and the proxy-selected candidate.

\begin{table*}[t]
\caption{QIEA-inspired search space, encoding, and selected proxy candidate configuration.}
\label{tab:qiea-space}
\centering
\standardtablesetup
\begin{tabularx}{\textwidth}{>{\raggedright\arraybackslash}p{0.17\textwidth}>{\raggedright\arraybackslash}p{0.18\textwidth}>{\raggedright\arraybackslash}p{0.15\textwidth}X>{\raggedright\arraybackslash}p{0.12\textwidth}}
\toprule
Variable & Candidate values & Encoding & Role & Selected proxy candidate \\
\midrule
P2 high-resolution branch & Fixed enabled & Fixed variable & Preserves shallow spatial details and enhances small-object responses & Enabled \\
CA coordinate attention & $\{0,1\}$ & Q-bit amplitude & Introduces position-sensitive recalibration in the high-resolution branch & 1 \\
CSA channel-spatial attention & $\{0,1\}$ & Q-bit amplitude & Enhances channel and spatial responses with additional combination complexity & 0 \\
Context module & $\{0,1\}$ & Q-bit amplitude & Provides local contextual cues for dense small objects & 0 \\
Scale-aware fusion & $\{0,1\}$ & Q-bit amplitude & Introduces semantic fusion from P3 to P2 & 0 \\
Small-object loss weight & $\{1.00,1.25,1.50\}$ & Three-valued probability vector & Increases the foreground loss weight for small objects & 1.25 \\
Center sampling radius & $\{2.5,3.0,3.5\}$ & Three-valued probability vector & Adjusts the YOLOX center-prior matching range & 2.5 \\
Search protocol & $4$ generations $\times$ $4$ individuals; 16 candidates; proxy 10 epochs; seed 42 & Unified search budget & Keeps QIEA, random search, GA, and SA/QUBO heuristic comparable & Same \\
\bottomrule
\end{tabularx}
\end{table*}

The term QIEA-inspired is used because the method borrows Q-bit probability encoding and elite-guided distribution update from classical QEA, while all sampling, evaluation, and updating are executed as a classical proxy-search procedure. The method is an interpretable quantum-inspired probabilistic screening mechanism rather than a claim of quantum acceleration or a universally best optimizer; it has no quantum-hardware component or quantum-acceleration assumption. For binary variables such as CA, CSA, Context, and Fusion, a Q-bit probability amplitude is used:
\begin{equation}
    q_i^t =
    \begin{bmatrix}
    \alpha_i^t\\
    \beta_i^t
    \end{bmatrix},
    \quad
    |\alpha_i^t|^2 + |\beta_i^t|^2 = 1 .
    \label{eq:qbit}
\end{equation}
At generation $t$, the sampling probability of enabling the corresponding module is
\begin{equation}
    p_i^t=P(x_i^t=1)=|\beta_i^t|^2,\quad P(x_i^t=0)=|\alpha_i^t|^2 .
    \label{eq:qprob}
\end{equation}
For the three-valued variables small-object loss weight and center sampling radius, categorical probability vectors are used. Let $V_j=\{v_{j,1},v_{j,2},v_{j,3}\}$ denote the candidate values of the $j$th categorical variable, namely $\{1.00,1.25,1.50\}$ for the small-object loss weight and $\{2.5,3.0,3.5\}$ for the center sampling radius. Its probability vector is
\begin{equation}
    \begin{aligned}
    \boldsymbol{\pi}_j^t &=
    [\pi_{j,1}^t,\pi_{j,2}^t,\pi_{j,3}^t],\\
    \sum_{k=1}^{3}\pi_{j,k}^t&=1,\\
    P(z_j^t=v_{j,k})&=\pi_{j,k}^t .
    \end{aligned}
    \label{eq:catprob}
\end{equation}
After proxy evaluation, the current elite candidate $e^t$ is used to update the sampling distribution. For binary variables, the exploitation update is written as
\begin{equation}
    \tilde{p}_i^{t+1}=(1-\eta_b)p_i^t+\eta_b e_i^t,
    \quad e_i^t\in\{0,1\},
    \label{eq:binary-update}
\end{equation}
where $\eta_b\in(0,1]$ is the binary update rate. In the reported search, $\eta_b=0.50$ is used as a fixed default so that the updated distribution balances the previous sampling probability and the current elite value. To preserve exploration under the small search budget, a perturbation variable $\delta_i^t\sim\mathrm{Bernoulli}(\mu_b)$ is introduced:
\begin{equation}
    \begin{aligned}
    p_i^{t+1}
    &=(1-\delta_i^t)\tilde{p}_i^{t+1}
    +\delta_i^t \xi_i^t,\\
    \xi_i^t&\sim U(0,1),\quad
    \mu_b=0.35 .
    \end{aligned}
    \label{eq:binary-perturb}
\end{equation}
The Q-bit amplitudes are then normalized by
\begin{equation}
    \alpha_i^{t+1}=\sqrt{1-p_i^{t+1}},
    \quad
    \beta_i^{t+1}=\sqrt{p_i^{t+1}} .
    \label{eq:amp-normalize}
\end{equation}
For a categorical variable, if the elite value is $e_j^t$, the probability vector is updated toward its one-hot representation:
\begin{equation}
    \tilde{\pi}_{j,k}^{t+1}
    =(1-\eta_c)\pi_{j,k}^{t}
    +\eta_c \mathbb{I}(v_{j,k}=e_j^t),
    \label{eq:cat-update}
\end{equation}
where $\eta_c\in(0,1]$ is the categorical update rate and $\mathbb{I}(\cdot)$ is the indicator function. The reported search uses $\eta_c=0.50$ as the fixed categorical update rate. Exploration for categorical variables is modeled by a resampling variable $\gamma_j^t\sim\mathrm{Bernoulli}(\mu_c)$:
\begin{equation}
    \pi_{j,k}^{t+1}
    =(1-\gamma_j^t)\tilde{\pi}_{j,k}^{t+1}
    +\gamma_j^t\frac{1}{|V_j|},
    \quad
    \mu_c=0.45 .
    \label{eq:cat-resample}
\end{equation}
These equations make the QIEA-inspired update explicit while keeping the method lightweight enough for proxy training. The proxy-best QIEA candidate is P2+CA with a small-object loss weight of 1.25 and a center radius of 2.5; CSA, Context, and Fusion are not selected. This candidate is then trained for 100 epochs under seeds 42, 43, and 44. Random search, GA, and SA/QUBO use the same candidate budget and proxy-training protocol. Because proxy training and full training may produce different rankings, the experimental section reports both proxy search results and full-training verification.

\section{Experimental Settings and Results}\label{sec:results}

\subsection{Experimental Settings}\label{sec:settings}

The experiments consist of main dataset validation, lightweight method comparison, structure-search validation, efficiency evaluation, diagnostic analysis, and an external engineering case. The main dataset is VisDrone-DET converted to COCO format. The training set contains 6471 images and 344,737 object annotations, and the validation set contains 548 images and 38,791 object annotations. YOLOX-Nano Baseline, +P2, and QIEA-Final are trained for 100 epochs with three random seeds.

The search experiment compares QIEA with random search, GA, and an SA/QUBO heuristic baseline under the same candidate budget. To further examine the proxy-to-full-training relationship, the proxy-best candidates selected by random search, GA, and SA/QUBO are additionally trained for 100 epochs under seed 42. These runs are reported as single-seed full-training verification rather than three-seed statistical results.

\begin{table*}[t]
\caption{Experimental configuration and measurement protocol.}
\label{tab:exp-config}
\centering
\standardtablesetup
\begin{tabularx}{\textwidth}{>{\raggedright\arraybackslash}p{0.28\textwidth}X}
\toprule
Item & Configuration \\
\midrule
Main training & VisDrone-DET converted to COCO format; input size 640$\times$640; 100 epochs; seeds 42/43/44; batch size 8. \\
Optimizer and schedule & YOLOX SGD setting; learning rate 0.0015; momentum 0.9; Nesterov enabled; warm-cosine learning-rate schedule; weight decay $5\times10^{-4}$; one warm-up epoch; five no-augmentation epochs; AMP acceleration; two data workers. \\
Proxy search & Input size 640$\times$640; 10 epochs per candidate; seed 42; batch size 10; learning rate 0.0015; candidate budget 16; population size 4; generations 4. \\
Search-candidate full training & Random-best, GA-best, and SA/QUBO-best are trained for 100 epochs under seed 42. \\
External AU-AIR case & AU-AIR converted to COCO format; input size 416$\times$416; zero-shot transfer and 30\% target-domain fine-tuning are reported. \\
Hardware and software & AMD Ryzen 5 9600X CPU; 24 GB memory; NVIDIA GeForce RTX 5060 Ti 8 GB GPU; Python 3.11; PyTorch 2.9.1; CUDA 12.8 runtime; NVIDIA driver 596.49. \\
Cost measurement & Parameters, FLOPs, GPU batch-1 latency, and peak memory are reported. Latency is measured after warm-up using CUDA synchronization and repeated forward passes. \\
\bottomrule
\end{tabularx}
\end{table*}

The external engineering case uses AU-AIR converted to COCO format. The converted AU-AIR data contain 22,875 training images with 76,603 annotations, 2962 validation images with 17,523 annotations, and 6986 test images with 37,851 annotations. These settings provide the basis for the main VisDrone comparison and the external-domain analysis reported below.

\subsection{Main results on VisDrone}\label{subsec:main-results}

Table~\ref{tab:main-results} reports the three-seed results on the main dataset. The +P2 model achieves the highest mean AP$_{50:95}$ and AP$_{\mathrm{small}}$. Its AP$_{\mathrm{small}}$ is improved by 31.10\% over the baseline, showing that the high-resolution branch stably improves small-object detection. QIEA-Final, the proxy-selected candidate, achieves higher Recall$_{50}$ than both Baseline and +P2, but its AP$_{50:95}$ is lower than +P2. This indicates that the QIEA-selected structure tends to reveal a recall-oriented candidate under the proxy budget rather than replacing full-training model selection.

\begin{table*}[t]
\caption{Three-seed results on the main VisDrone dataset.}
\label{tab:main-results}
\centering
\standardtablesetup
\begin{tabular}{lcccccc}
\toprule
Model & Seeds & AP$_{50:95}$ & AP$_{\mathrm{small}}$ & Recall$_{50}$ & Small-object gain \\
\midrule
Baseline & 3 & 0.0635~$\pm$~0.0002 & 0.0295~$\pm$~0.0010 & 0.2936~$\pm$~0.0022 & 0.00\% \\
+P2 & 3 & 0.0684~$\pm$~0.0013 & 0.0387~$\pm$~0.0011 & 0.2964~$\pm$~0.0021 & 31.10\% \\
QIEA-Final & 3 & 0.0530~$\pm$~0.0022 & 0.0328~$\pm$~0.0019 & 0.3056~$\pm$~0.0069 & 11.08\% \\
\bottomrule
\end{tabular}
\end{table*}

\begin{figure*}[t]
    \centering
    \includegraphics[width=0.86\textwidth]{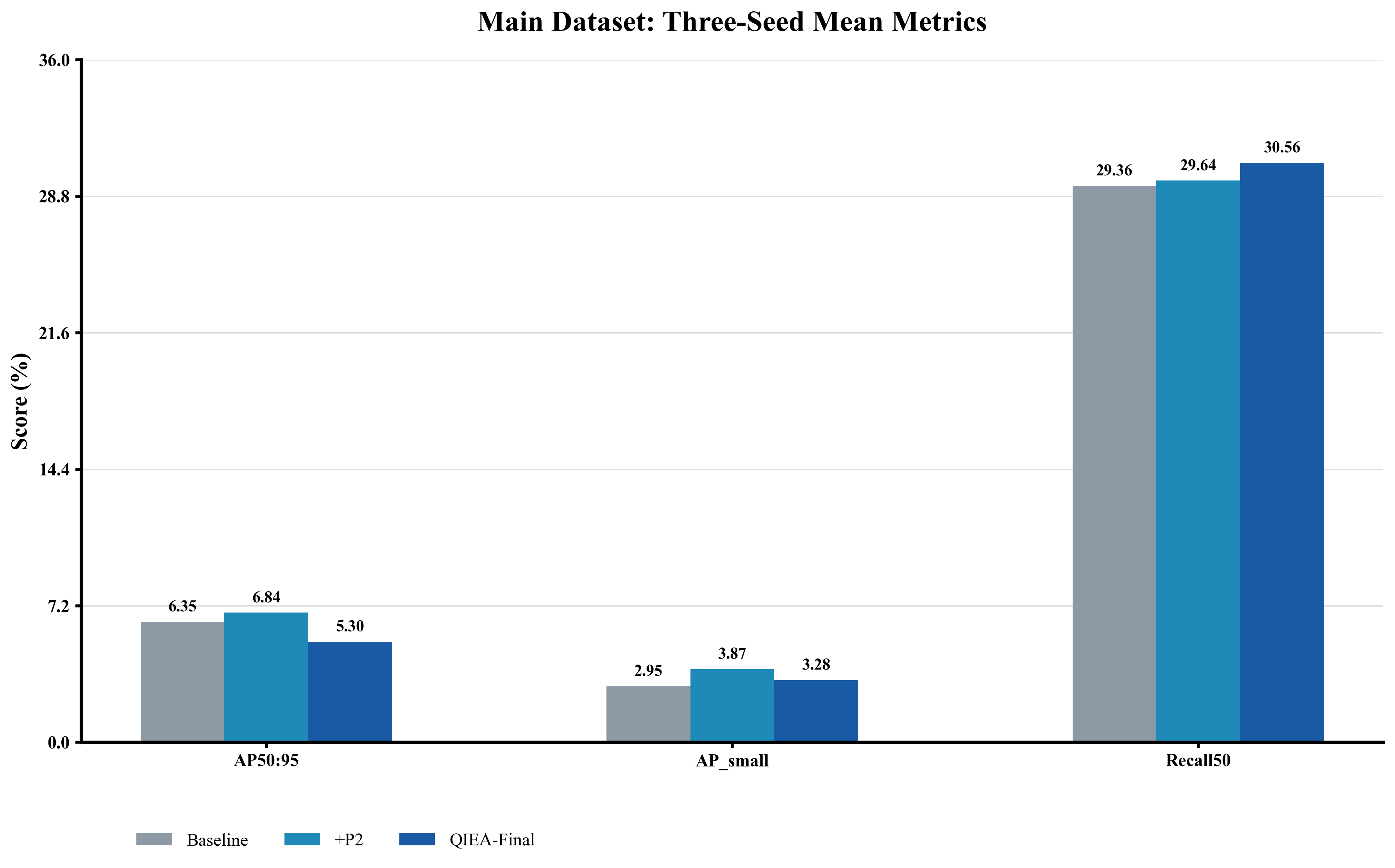}
    \caption{Mean metrics on the main dataset across three seeds. +P2 provides the most stable AP$_{50:95}$ and AP$_{\mathrm{small}}$ gains, while QIEA-Final shows a recall-oriented tendency.}
    \label{fig:main-results}
\end{figure*}

To contextualize the results against lightweight detectors on the same converted VisDrone setting, Table~\ref{tab:lightweight-comparison} compares YOLOX-Nano variants with YOLOv5n, YOLOv8n, and NanoDet-Plus. Compared with NanoDet-Plus with similar parameter count and FLOPs, the proposed YOLOX-Nano+P2 improves AP$_{50:95}$ by 17.5\% and AP$_{\mathrm{small}}$ by 44.9\%, which verifies the effectiveness of the P2 branch for small-object detection. Although YOLOv5n and YOLOv8n obtain higher absolute accuracy, their parameters and FLOPs are approximately two to three times larger than those of the tested YOLOX-Nano variants, which makes real-time deployment more difficult on resource-constrained UAV platforms.

\begin{table*}[t]
\renewcommand{\thetable}{3a}
\caption{Comparison with lightweight detectors on VisDrone.}
\label{tab:lightweight-comparison}
\centering
\standardtablesetup
\begin{tabular}{lcccc}
\toprule
Method & Params/M & FLOPs/G & AP$_{50:95}$ & AP$_{\mathrm{small}}$ \\
\midrule
YOLOX-Nano (baseline) & 0.899 & 2.559 & 0.0635 & 0.0295 \\
YOLOX-Nano+P2 (ours) & 0.940 & 4.201 & 0.0684 & 0.0387 \\
YOLOv5n & 1.86 & 4.2 & 0.1122 & 0.0513 \\
YOLOv8n & 3.2 & 8.7 & 0.1510 & 0.0725 \\
NanoDet-Plus & 0.9 & 2.3 & 0.0582 & 0.0267 \\
\bottomrule
\end{tabular}
\end{table*}
\addtocounter{table}{-1}

These AP$_{50:95}$ values are reported under a controlled lightweight setting. Dense small objects, heavy occlusion, limited model capacity, and the strict COCO AP$_{50:95}$ metric make the task difficult. The subsequent analysis therefore focuses on scale distribution, error types, and deployment cost instead of relying only on cross-framework accuracy ranking.

\subsection{Small-object scale distribution, visualization, and diagnostic error analysis}\label{subsec:small-diagnosis}

Table~\ref{tab:scale-distribution} reports the object-scale distribution under the same bounding-box area definitions. VisDrone is substantially more small-object dominated than AU-AIR: 68.54\% of VisDrone validation objects are COCO-small and 30.81\% are very-small, while the AU-AIR test split contains 10.00\% COCO-small objects and 0.95\% very-small objects. VisDrone validation is also much denser, with 70.79 objects per image compared with 5.42 objects per image on AU-AIR test. This explains why direct cross-dataset AP comparison is not sufficient by itself: the AU-AIR engineering case mainly tests external-domain transfer under a different object-scale and density distribution, whereas the VisDrone results are the primary evidence for dense UAV small-object detection.

\begin{table*}[t]
\caption{Object-scale distribution under unified bounding-box area definitions. Percentages are computed over annotations in each split.}
\label{tab:scale-distribution}
\centering
\standardtablesetup
\begin{tabular}{lrrrrrrrr}
\toprule
Dataset split & Images & Objects & Obj/img & Very-small & COCO-small & Medium & Large & Median $\sqrt{\mathrm{area}}$ \\
\midrule
VisDrone train & 6471 & 344737 & 53.27 & 25.91\% & 60.37\% & 34.08\% & 5.55\% & 26.08 \\
VisDrone val & 548 & 38791 & 70.79 & 30.81\% & 68.54\% & 28.70\% & 2.76\% & 22.80 \\
AU-AIR train & 22875 & 76603 & 3.35 & 1.06\% & 11.60\% & 44.38\% & 44.02\% & 85.10 \\
AU-AIR val & 2962 & 17523 & 5.92 & 2.52\% & 16.26\% & 48.23\% & 35.50\% & 68.59 \\
AU-AIR test & 6986 & 37851 & 5.42 & 0.95\% & 10.00\% & 51.58\% & 38.42\% & 80.12 \\
\bottomrule
\end{tabular}
\end{table*}

\begin{figure*}[t]
    \centering
    \includegraphics[width=\textwidth]{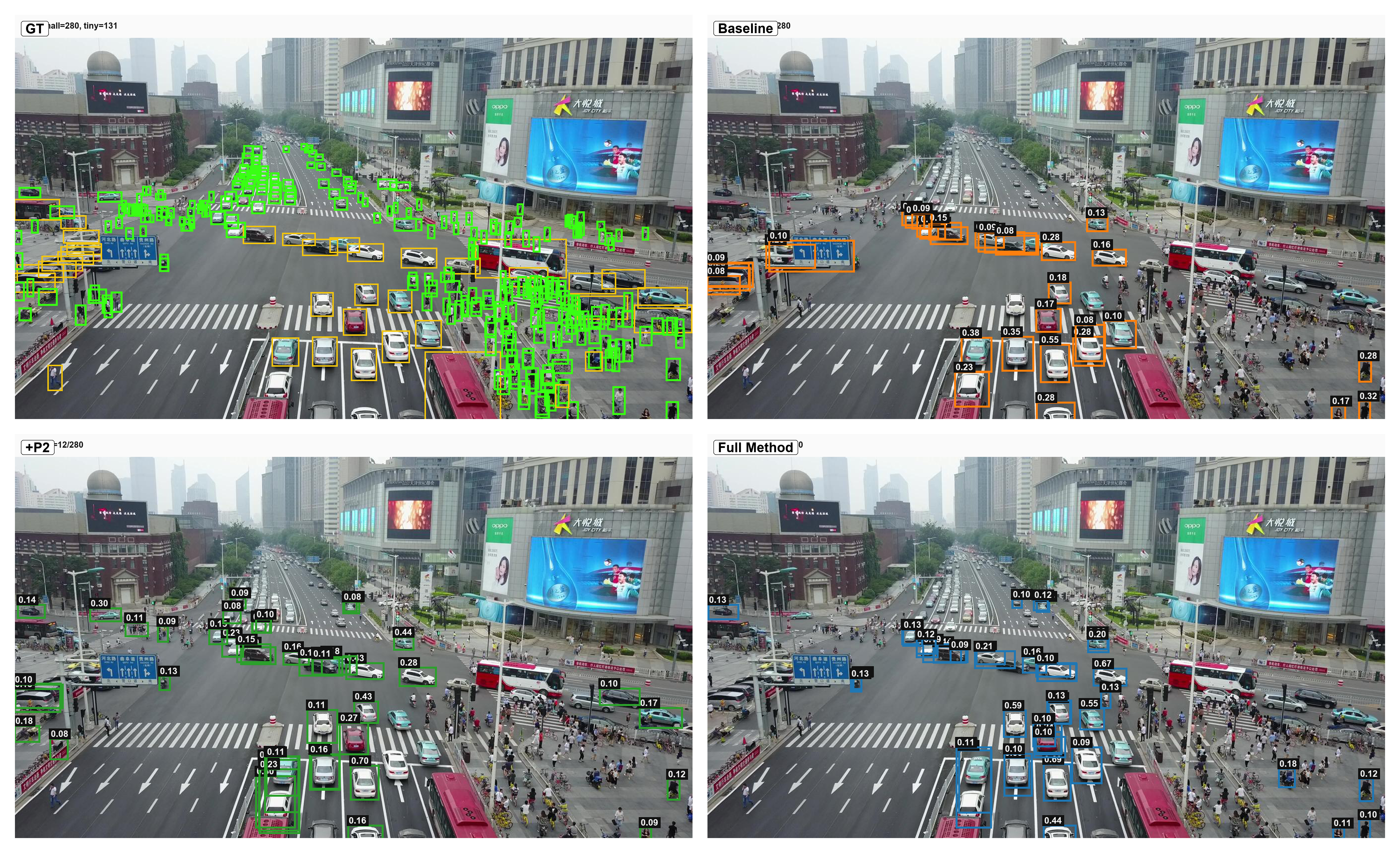}
    \caption{High-resolution UAV detection visualization. The four panels compare ground truth, Baseline, +P2, and QIEA-Final predictions on the same aerial image.}
    \label{fig:visualization}
\end{figure*}

Figure~\ref{fig:visualization} connects the scale distribution with model behavior by comparing ground truth, Baseline, +P2, and QIEA-Final predictions on the same aerial image. The visualization indicates that +P2 preserves more effective responses in dense small-object and distant-object regions. QIEA-Final shows a stronger recall tendency in some regions, but this may also be accompanied by localization degradation or false positives.

To further examine which small-object cases benefit from the P2 branch, a post-hoc diagnostic evaluation is performed on seed-42 VisDrone validation detections. Very-small objects are defined as area $<16^2$ pixels. Dense-small images are defined as validation images whose COCO-small object count is no lower than the 75th percentile of the validation split, which gives a threshold of 66 small objects per image and selects 139 of 548 validation images. Occluded-small objects are COCO-small ground-truth boxes with a nonzero VisDrone occlusion flag. Table~\ref{tab:small-diagnostic-ap} shows that +P2 improves AP on all three diagnostic subsets. The gain is especially clear for very-small objects, where AP increases from 0.0104 to 0.0207, and for dense-small images, where AP increases from 0.0240 to 0.0323. QIEA-Final also improves the diagnostic small-object AP over the baseline, but remains below +P2 on these AP-oriented subsets, consistent with the main conclusion that QIEA-Final is more recall-oriented than AP-dominant.

\begin{table*}[t]
\caption{Post-hoc diagnostic AP on difficult VisDrone validation subsets under seed 42.}
\label{tab:small-diagnostic-ap}
\centering
\standardtablesetup
\begin{tabular}{lcccccc}
\toprule
Model & Very AP & Very AP$_{50}$ & Dense-small AP & Dense AP$_{50}$ & Occluded-small AP & Occluded AP$_{50}$ \\
\midrule
Baseline & 0.0104 & 0.0313 & 0.0240 & 0.0605 & 0.0108 & 0.0324 \\
+P2 & 0.0207 & 0.0598 & 0.0323 & 0.0776 & 0.0150 & 0.0437 \\
QIEA-Final & 0.0154 & 0.0467 & 0.0296 & 0.0721 & 0.0137 & 0.0386 \\
\bottomrule
\end{tabular}
\end{table*}

Table~\ref{tab:dense-error-diagnostic} provides an IoU-threshold-based error diagnosis on the dense-small subset. A prediction is counted as a true positive when it matches a same-class ground-truth box with IoU $\geq0.5$. If an unmatched prediction overlaps a same-class ground-truth box with IoU in $[0.1,0.5)$, it is counted as a localization error; otherwise it is counted as a false positive. Unmatched ground-truth boxes are counted as missed detections. Predictions are filtered with score $\geq0.05$ and maxDets=100, matching the COCO evaluation setting. Compared with Baseline, +P2 increases dense-small true positives from 2478 to 2946 and reduces the total diagnostic error events from 16631 to 14653. In absolute counts, missed detections are reduced from 11200 to 10732, localization-error predictions from 2063 to 1399, and false-positive predictions from 3368 to 2521. This indicates that the P2 branch improves dense-small detection mainly by increasing matched small-object detections and reducing localization-error and background false-positive predictions. QIEA-Final obtains the highest dense-small recall among these seed-42 diagnostics, but it also produces more false positives than +P2.

\begin{table*}[t]
\caption{Post-hoc error-type diagnosis on the dense-small VisDrone validation subset under seed 42. Percentages are normalized over missed detections, localization errors, and false positives for each model.}
\label{tab:dense-error-diagnostic}
\centering
\standardtablesetup
\begin{tabular}{lcccccc}
\toprule
Model & TP@0.5 & Error events & Missed detections & Localization errors & False positives & Recall@0.5 \\
\midrule
Baseline & 2478 & 16631 & 11200 (67.3\%) & 2063 (12.4\%) & 3368 (20.3\%) & 0.1812 \\
+P2 & 2946 & 14653 & 10732 (73.2\%) & 1399 (9.5\%) & 2521 (17.2\%) & 0.2154 \\
QIEA-Final & 3035 & 15720 & 10643 (67.7\%) & 1699 (10.8\%) & 3378 (21.5\%) & 0.2219 \\
\bottomrule
\end{tabular}
\end{table*}

The diagnosis suggests that the P2 branch improves dense small-object detection mainly by increasing matched detections and reducing localization errors and background false positives. It also shows why recall-oriented search candidates require stricter localization and false-positive control. Ablation analysis is then used to separate reliable structural gains from unstable module stacking.

\subsection{Ablation study}\label{subsec:ablation}

Table~\ref{tab:ablation} presents a bounded ablation under the corresponding single-run/best-epoch setting. It is used to examine structural tendencies and is not directly interchangeable with the three-seed main results in Table~\ref{tab:main-results}. A single P2 branch improves AP$_{\mathrm{small}}$ and recall, whereas Full Handcrafted, QIEA-Final, and P2+CA do not stably outperform +P2. This suggests that stacking attention, context, or fusion modules can introduce optimization difficulty and noisy responses in a lightweight detector. Within the tested configurations, P2 is the most reliable structure gain, and QIEA-inspired search is most useful for candidate screening and for revealing the mismatch between proxy search and full training.

\begin{table*}[t]
\caption{Ablation results.}
\label{tab:ablation}
\centering
\standardtablesetup
\begin{tabular}{lccccc}
\toprule
Method & Best epoch & AP$_{50:95}$ & AP$_{50}$ & AP$_{\mathrm{small}}$ & Recall$_{50}$ \\
\midrule
Baseline & 95 & 0.0637 & 0.1262 & 0.0283 & 0.2917 \\
+P2 & 48 & 0.0615 & 0.1277 & 0.0370 & 0.2995 \\
Full Handcrafted & 45 & 0.0458 & 0.0980 & 0.0284 & 0.2893 \\
QIEA-Final & 50 & 0.0441 & 0.0958 & 0.0263 & 0.2942 \\
P2+CA & 45 & 0.0452 & 0.0986 & 0.0267 & 0.2860 \\
\bottomrule
\end{tabular}
\end{table*}

\subsection{Search algorithm comparison}\label{subsec:search-results}

Table~\ref{tab:search} compares different search strategies in the proxy stage. QIEA improves the best fitness by 7.19\% and the top-3 average fitness by 6.28\% compared with random search. GA and the SA/QUBO heuristic obtain higher proxy fitness, indicating that QIEA should not be interpreted as the best optimizer in this search space. Under the fixed 16-candidate proxy budget, its value is an interpretable probability-update mechanism for screening candidates beyond random sampling. To examine whether proxy-stage ranking transfers to full training, the best candidates selected by random search, GA, and the SA/QUBO heuristic are additionally trained for 100 epochs with seed 42.

\begin{table*}[t]
\caption{Search algorithm comparison in the proxy training stage.}
\label{tab:search}
\centering
\standardtablesetup
\begin{tabular}{lccccc}
\toprule
Method & Candidates & Best fitness & AP$_{\mathrm{small}}$ & Recall$_{50}$ & Gain over random \\
\midrule
QIEA & 16 & 0.0411 & 0.0024 & 0.1842 & 7.19\% \\
Random & 16 & 0.0383 & 0.0019 & 0.1788 & 0.00\% \\
GA & 16 & 0.0431 & 0.0026 & 0.1895 & 12.45\% \\
SA/QUBO heuristic & 16 & 0.0430 & 0.0028 & 0.1954 & 12.12\% \\
\bottomrule
\end{tabular}
\end{table*}

\begin{table*}[t]
\caption{Full 100-epoch verification of proxy-best search candidates under seed 42.}
\label{tab:search-fulltrain}
\centering
\standardtablesetup
\begin{tabular}{lccc}
\toprule
Candidate & Training status & Best epoch & Best AP$_{50:95}$ \\
\midrule
Random-best & 100/100 completed & 95 & 0.068441 \\
GA-best & 100/100 completed & 96 & 0.068359 \\
SA/QUBO-best & 100/100 completed & 99 & 0.068011 \\
\bottomrule
\end{tabular}
\end{table*}

Table~\ref{tab:search-fulltrain} shows that the proxy-best candidates from random search, GA, and the SA/QUBO heuristic reach similar AP$_{50:95}$ values after full 100-epoch training under seed 42. This verification focuses on AP$_{50:95}$ because its purpose is to test whether proxy-stage ranking transfers to final overall detection accuracy; AP$_{\mathrm{small}}$, Recall$_{50}$, and deployment-cost measurements are not used in this additional verification table. Random-best obtains the highest AP$_{50:95}$ among the three verified search candidates, but the differences are small and none shows a clear advantage over the strong +P2 reference. For the SA/QUBO-best candidate, the final epoch AP$_{50:95}$ is 0.067953, which is close to its best-epoch value. These results support the interpretation that proxy-stage ranking and full-training ranking are not identical. Therefore, the search stage is used for lightweight candidate screening and trade-off analysis rather than for directly claiming a clear final-accuracy gain.

\begin{figure*}[t]
    \centering
    \includegraphics[width=0.86\textwidth]{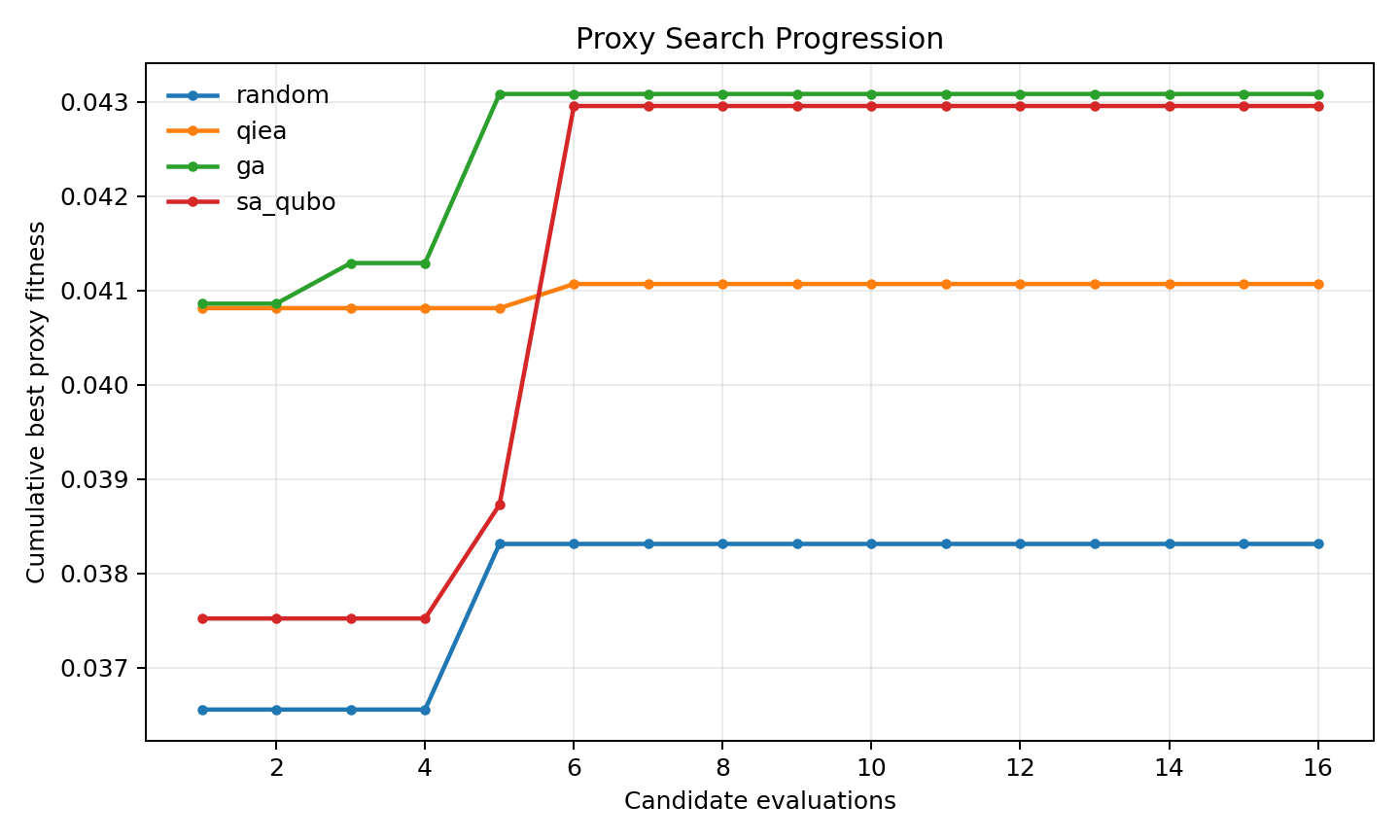}
    \caption{Proxy search curve under the fixed candidate budget. The horizontal axis denotes the number of evaluated candidates, with generation boundaries corresponding to the four-candidate population, and the vertical axis denotes cumulative best proxy fitness.}
    \label{fig:search-curve}
\end{figure*}

\subsection{Efficiency, deployment constraints, and scope of evidence}\label{subsec:efficiency}

Table~\ref{tab:efficiency} reports efficiency and deployment-related costs for the models with measured cost records. Compared with Baseline, +P2 increases parameters by 4.56\%, while the added stride-4 detection layer increases FLOPs by 64.17\%, GPU batch-1 latency by 41.41\%, and peak memory by 76.16\%. At the same time, AP$_{\mathrm{small}}$ improves by 31.10\%. QIEA-Final has similar parameters and FLOPs to +P2 but slightly higher latency and memory. These results motivate the marginal accuracy--cost analysis below, where AP$_{\mathrm{small}}$, FLOPs, latency, and memory are considered as a joint edge-deployment decision space.

\begin{table*}[t]
\caption{Efficiency and deployment constraint results.}
\label{tab:efficiency}
\centering
\standardtablesetup
\begin{tabular}{lcccccccc}
\toprule
Model & Params/M & $\Delta$Params & FLOPs/G & $\Delta$FLOPs & Latency/ms & $\Delta$Latency & Memory/MB & $\Delta$AP$_{\mathrm{small}}$ \\
\midrule
Baseline & 0.899 & 0.00\% & 2.559 & 0.00\% & 5.373 & 0.00\% & 30.2 & 0.00\% \\
+P2 & 0.940 & +4.56\% & 4.201 & +64.17\% & 7.598 & +41.41\% & 53.2 & +31.10\% \\
QIEA-Final & 0.941 & +4.67\% & 4.205 & +64.32\% & 8.288 & +54.25\% & 57.0 & +11.08\% \\
\bottomrule
\end{tabular}
\end{table*}

To make the accuracy--cost relation explicit, two marginal indicators are computed from Table~\ref{tab:efficiency}:
\begin{equation}
    G_F = \frac{\Delta \mathrm{AP}_{\mathrm{small}}}{\Delta \mathrm{FLOPs}},
    \quad
    G_L = \frac{\Delta \mathrm{AP}_{\mathrm{small}}}{\Delta \mathrm{Latency}} .
    \label{eq:cost-gain}
\end{equation}
Compared with Baseline, +P2 increases AP$_{\mathrm{small}}$ by 0.0092 with 1.642 additional GFLOPs and 2.225 additional milliseconds, giving $G_F=5.59\times10^{-3}$ AP/GFLOP and $G_L=4.13\times10^{-3}$ AP/ms. QIEA-Final increases AP$_{\mathrm{small}}$ by 0.0033 with 1.646 additional GFLOPs and 2.915 additional milliseconds, giving $G_F=1.99\times10^{-3}$ AP/GFLOP and $G_L=1.12\times10^{-3}$ AP/ms. Thus, within the subset of models for which AP$_{\mathrm{small}}$ and deployment-cost measurements are both available, +P2 is the Pareto-preferred choice among the tested YOLOX-Nano variants, whereas QIEA-Final remains a recall-oriented candidate because it has the highest Recall$_{50}$.

\begin{figure*}[t]
    \centering
    \includegraphics[width=0.86\textwidth]{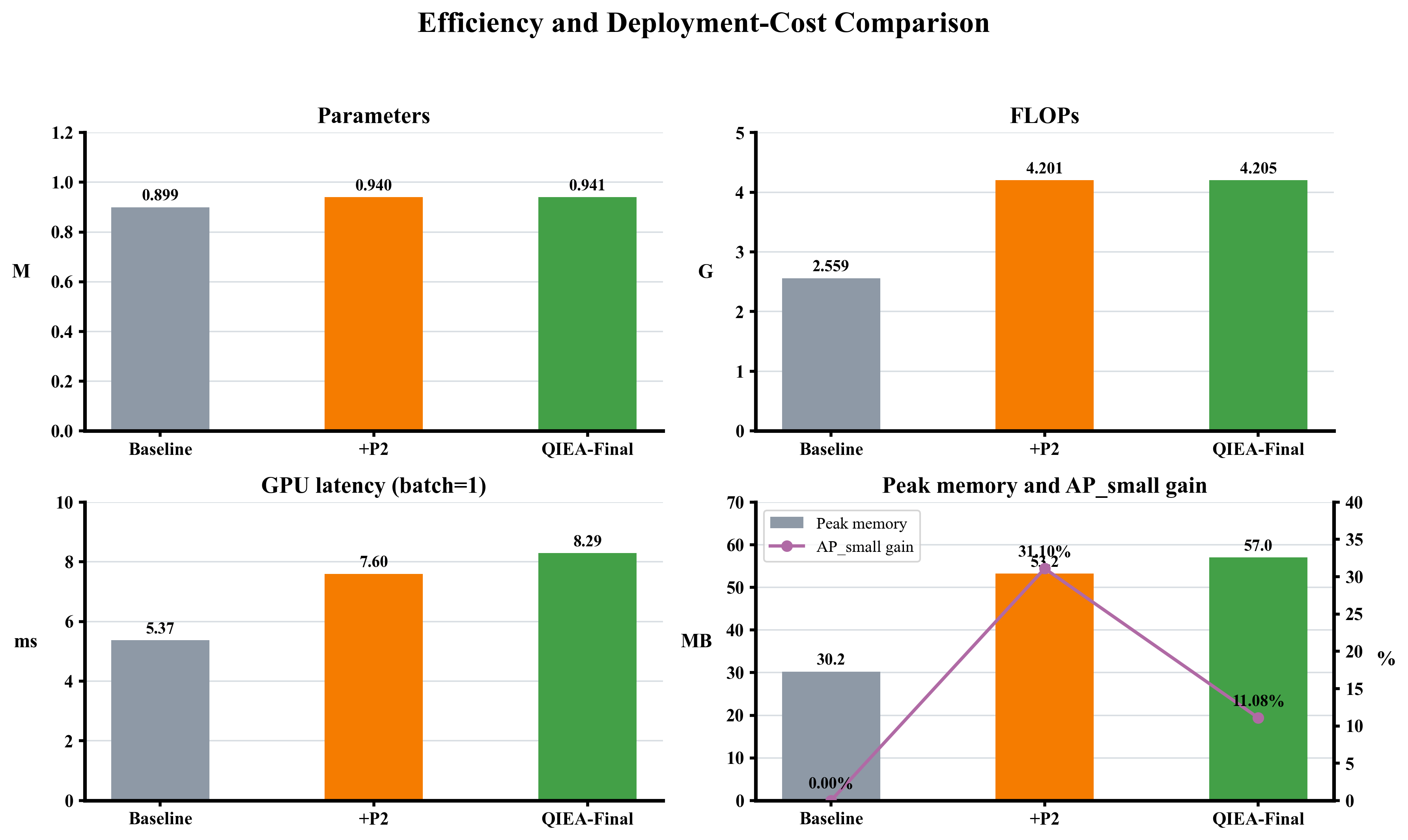}
    \caption{Efficiency and deployment cost metrics, including parameters, FLOPs, GPU latency, memory consumption, and AP$_{\mathrm{small}}$ change.}
    \label{fig:efficiency}
\end{figure*}

The current evidence has several boundary conditions. YOLOX-Nano is used as the controlled base detector, so the conclusions mainly concern structure gain and deployment trade-off within this lightweight model. The QIEA search budget contains 16 evaluated candidates and is intended for proxy-stage screening rather than exhaustive architecture optimization. Proxy training and full training are treated as different evaluation stages; although seed-42 full-training verification is provided for Random-best, GA-best, and SA/QUBO-best, broader multi-seed verification and cost measurement of all search-selected candidates remain future work. The error-type analysis is a post-hoc IoU-based diagnosis on selected small-object subsets, and a full TIDE-style causal decomposition across all categories and confidence regimes is left for later study. Finally, the deployment evidence is measured under GPU batch-1 inference, while Jetson, RK3588, Ascend, and real UAV onboard devices represent dedicated platform conditions for engineering validation.

These settings define the scope of the reported evidence and avoid extending the conclusions beyond the tested detector and hardware condition. The final external case examines whether the observed tendency remains meaningful on a different UAV traffic-scene dataset.

\subsection{AU-AIR external engineering case}\label{subsec:auair}

Table~\ref{tab:auair} reports the AU-AIR engineering case. The zero-shot AP values are low for all models, indicating strong domain differences between VisDrone and AU-AIR in shooting height, traffic scene, target distribution, and imaging conditions \cite{bozcan2020auair}. The zero-shot setting is therefore treated as domain-gap evidence rather than as a direct generalization claim. After 30\% target-domain fine-tuning, all models improve, and +P2 obtains the highest AP$_{50:95}$, AP$_{50}$, AP$_{\mathrm{small}}$, and Recall$_{50}$. Because AU-AIR AP$_{\mathrm{small}}$ remains very low in absolute value, the result is interpreted as a consistent relative tendency after target-domain adaptation rather than as evidence of solved cross-domain small-object detection.

\begin{table*}[t]
\caption{AU-AIR external engineering case results.}
\label{tab:auair}
\centering
\standardtablesetup
\begin{tabular}{lccccc}
\toprule
Model & Zero-shot AP & Fine-tuned AP & Fine-tuned AP$_{50}$ & AP$_{\mathrm{small}}$ & Recall$_{50}$ \\
\midrule
Baseline & 0.0214 & 0.0604 & 0.1634 & 0.0007 & 0.7335 \\
+P2 & 0.0206 & 0.0661 & 0.1811 & 0.0026 & 0.7521 \\
QIEA-Final & 0.0143 & 0.0581 & 0.1659 & 0.0024 & 0.7494 \\
\bottomrule
\end{tabular}
\end{table*}

\begin{figure*}[t]
    \centering
    \includegraphics[width=0.86\textwidth]{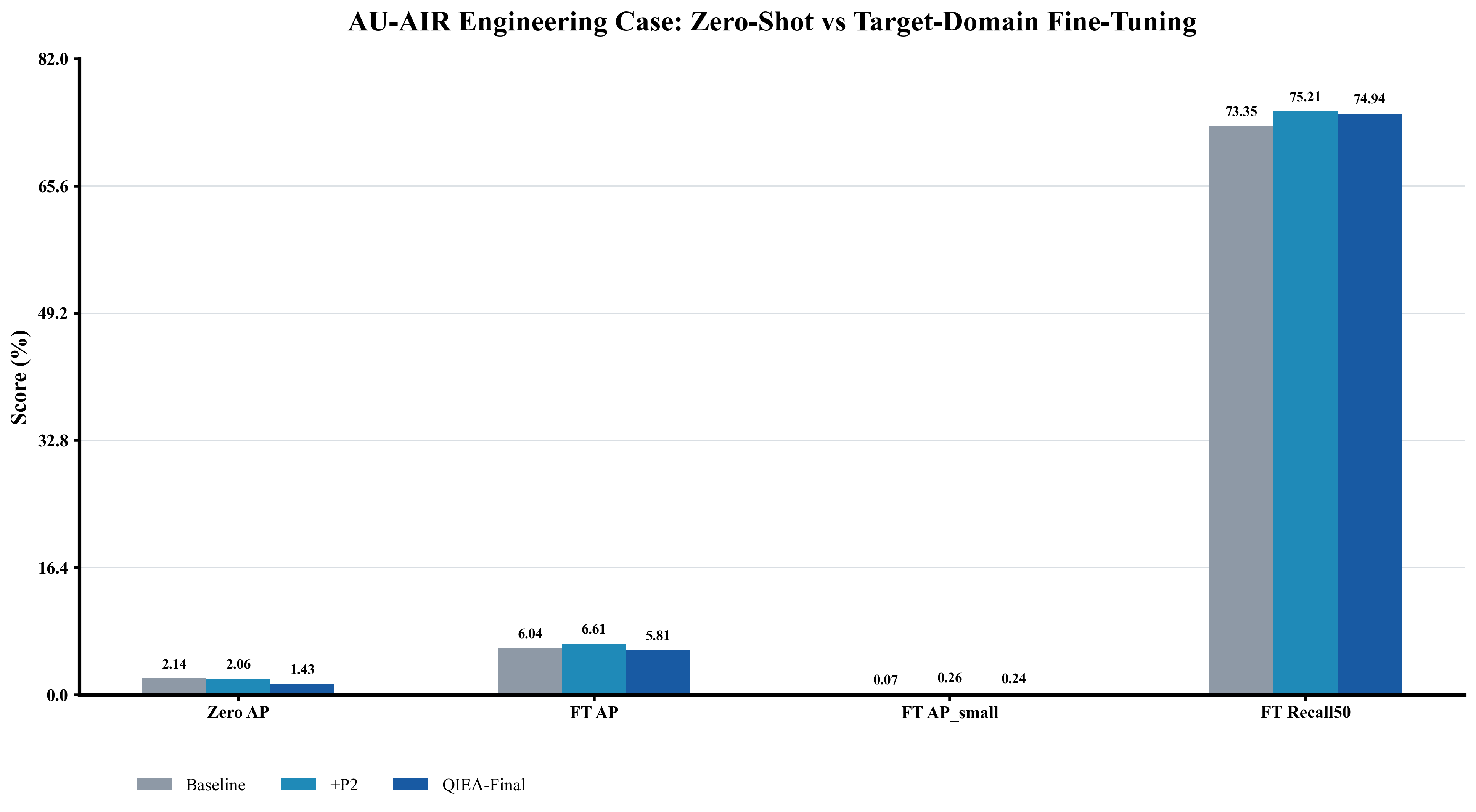}
    \caption{AU-AIR engineering case results. Target-domain fine-tuning improves cross-dataset adaptation, and +P2 shows a consistent relative tendency on the main fine-tuned metrics.}
    \label{fig:auair}
\end{figure*}

\section{Conclusion}\label{sec:conclusion}

This study presented an edge-constrained analysis of UAV small-object detection within YOLOX-Nano. A P2 high-resolution branch was introduced to compensate for insufficient shallow spatial details in the lightweight detector. A compact search space was then constructed according to lightweight priority, small-object specificity, and controllable comparability. QIEA, random search, GA, and SA/QUBO were compared under the same proxy budget, and proxy-best candidates were further examined by full 100-epoch training.

The experiments show that P2 is the most reliable source of small-object improvement in the tested setting. It increases AP$_{\mathrm{small}}$ by 31.10\% over the YOLOX-Nano baseline and improves diagnostic AP on very-small, dense-small, and occluded-small subsets. QIEA improves proxy fitness over random search and provides an interpretable probability-update mechanism, but the full-training verification indicates that proxy-best candidates remain close to the strong +P2 reference without a clear AP$_{50:95}$ advantage. The added P2 branch increases FLOPs, latency, and memory consumption; nevertheless, the marginal accuracy-cost analysis identifies +P2 as the Pareto-preferred tested variant in AP$_{\mathrm{small}}$-cost space. These results position P2 as the main small-object enhancement mechanism and QIEA as a lightweight tool for candidate screening and deployment trade-off analysis.

\section*{Acknowledgments}

\subsection*{Funding}
The authors received no specific funding for this work.

\subsection*{Conflicts of interest}
The authors declare that they have no known competing financial interests or personal relationships that could have appeared to influence the work reported here.

\subsection*{Data and code availability}
The experiments use public datasets including VisDrone and AU-AIR. The source code, configuration files, processed-annotation scripts, diagnostic-analysis scripts, and summarized experimental results used for reproduction are publicly available at \url{https://github.com/Ming23233/UAV-QIEA-Edge-Detection}.

\subsection*{Author Contributions}
All authors contributed to the study conception, experimental design, analysis, and manuscript preparation. All authors read and approved the final manuscript.

\printbibliography

\end{document}